%% file: main.tex
\documentclass[10pt,twocolumn,letterpaper]{article}
\usepackage[T1]{fontenc}   % must be loaded once
\usepackage[utf8]{inputenc} % optional but fine with pdfLaTeX

\usepackage{wacv}              % To produce the CAMERA-READY version
\input{preamble}

\definecolor{wacvblue}{rgb}{0.21,0.49,0.74}
\usepackage[pagebackref,breaklinks,colorlinks,allcolors=wacvblue]{hyperref}
\usepackage[accsupp]{axessibility}  % Improves PDF readability for those with disabilities.

\usepackage{xcolor}

\def\wacvPaperID{1654} % *** Enter the WACV Paper ID here
\def\confName{WACV}
\def\confYear{2026}

\newcommand{\SYS}{mmWEAVER\xspace}
\newcommand{\sys}{mmWeaver\xspace}
\title{\SYS: Environment-Specific mmWave Signal Synthesis from a Photo and Activity Description}

\author{Mahathir Monjur\\
UNC Chapel Hill\\
{\tt\small mahathir@cs.unc.edu} 
\and
Shahriar Nirjon\\
UNC Chapel Hill\\
{\tt\small nirjon@cs.unc.edu}
}

\usepackage{balance}
\begin{document}
%\maketitle

\twocolumn[{
    \maketitle
    \vspace{-2mm}
    \begin{center}
    \includegraphics[width=0.85\textwidth]{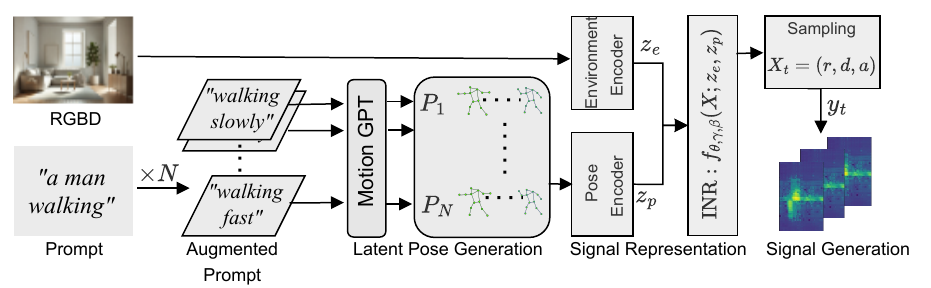}
    \captionof{figure}{System overview of \sys. Given an RGB-D image and a textual activity description, MotionGPT generates human pose sequence. The environment encoder extracts static scene features (\(z_e\)) from RGB, depth, and segmentation maps, while the pose encoder processes the pose sequence to produce temporally-aware motion features (\(\{z_p\}_t\)). These are fused and input into a hypernetwork that generates the weights (\(\theta\)) and time-based modulation parameters (\(\gamma(t), \beta(t)\)) for an Implicit Neural Representation (INR), which models the mmWave signal as a continuous function. The INR is then sampled over spatial-temporal coordinates to synthesize high-resolution mmWave signals tailored to the environment and motion context.}
    \label{fig:overview}
    \end{center}
    \vspace{2mm}
}]

\begin{abstract}
Realistic signal generation and dataset augmentation are essential for advancing mmWave radar applications such as activity recognition and pose estimation, which rely heavily on diverse, and environment-specific signal datasets. However, mmWave signals are inherently complex, sparse, and high-dimensional, making physical simulation computationally expensive. This paper presents \sys, a novel framework that synthesizes realistic, environment-specific complex mmWave signals by modeling them as continuous functions using Implicit Neural Representations (INRs), achieving up to 49-fold compression. \sys incorporates hypernetworks that dynamically generate INR parameters based on environmental context (extracted from RGB-D images) and human motion features (derived from text-to-pose generation via MotionGPT), enabling efficient and adaptive signal synthesis. By conditioning on these semantic and geometric priors, \sys generates diverse I/Q signals at multiple resolutions, preserving phase information critical for downstream tasks such as point cloud estimation and activity classification. Extensive experiments show that \sys achieves a complex SSIM of 0.88 and a PSNR of 35 dB, outperforming existing methods in signal realism while improving activity recognition accuracy by up to 7\% and reducing human pose estimation error by up to 15\%, all while operating $6$–$35\times$ faster than simulation-based approaches.
\end{abstract}
%% Abstract %%

\input{sections_revised/01_introduction}
\input{sections_revised/02_related}
\input{sections_revised/04_method}
\input{sections_revised/05_microbenchmark}
\input{sections_revised/06_evaluation}
\input{sections_revised/07_conclusion}

\section*{Acknowledgments}
This work was supported in part by the National Science Foundation under NSF Awards 2503073 and 2047461.

\balance
\small
\bibliographystyle{ieeenat_fullname}
\bibliography{references}

\end{document}

%% file: preamble.tex
\newcommand{\parlabel}[1]{\par\noindent\textbf{#1} }
\usepackage{comment}
\usepackage{caption}
\usepackage{subcaption}
\usepackage{pbalance}

%% file: sections_revised/01_introduction.tex
\section{Introduction}
\label{sec:intro}

mmWave sensing is a promising technology for indoor applications such as patient monitoring~\cite{patientmonitoring1,patientmonitoring2,patientmonitoring3}, activity recognition~\cite{har1,har2}, and pose estimation~\cite{pose1,hupr,pose3, pose4}, offering key advantages over traditional vision systems. Unlike optical cameras, mmWave radars preserve privacy by avoiding detailed imaging, making them ideal for sensitive environments like homes and healthcare settings. They function well in low-light conditions and can penetrate clothing and some walls, enabling robust motion tracking. However, factors such as environmental sensitivity, room layout variation, material properties and deployment cost pose significant challenges to widespread adoption.

Synthetic mmWave data generation has emerged as a promising solution to dataset scarcity, offering a rapid and cost-effective alternative to large-scale data collection. Techniques such as data augmentation through transformations~\cite{wang2023data}, RF simulations~\cite{sim1,wang2023data,rf_genesis}, and model-based generative methods~\cite{xue2023towards,huang2023deep,zhang2022synthesized,rfdiffusion} enhance dataset diversity and improve model robustness. However, generative models often fail to incorporate environment-specific details, overlooking critical propagation effects unique to indoor settings. Simulation-based methods, though reliable in idealized empty environments, are computationally expensive and may not reflect the noise and variability of real-world data.

When the deployment environment is known, we advocate for environment-specific synthetic data generation, which offers greater accuracy than generic augmentation. Effective tools for this task should satisfy three criteria: (i) generate realistic data that capture both human motion and environment-specific characteristics, (ii) require minimal yet informative inputs encoding scene and activity context, and (iii) produce mmWave signals in a storage-efficient, shareable, and resolution-scalable format suitable for diverse applications.

In this paper, we introduce \textit{\sys}, a framework that generates environment-specific mmWave signals for diverse human activities from two simple inputs: a photo of the room (capturing geometry, objects, and materials) and a textual activity description (e.g., \textit{``user seated on couch drinking water''}). \sys synthesizes realistic signals that reflect human gestures and movements in the target environment, employs a space-efficient representation to address storage and sharing limits, and supports multi-resolution sampling via parameterized data representations. It also avoids overfitting to minor environmental changes and enables large-scale dataset creation through unlimited input-output variations. By reducing costly real-world data collection, \sys facilitates reuse of synthetic data across diverse sensing and AI applications.

Modeling mmWave signal propagation in indoor environments is challenging due to complex interactions between static factors (e.g., multipath from walls and furniture) and dynamic factors (e.g., body shape, pose, and motion). To address this, we propose \sys, which leverages Implicit Neural Representations (INRs)~\cite{inr} to model signals as continuous, differentiable functions, enabling multi-resolution synthesis and compact storage. To adapt across scenarios, \sys uses hypernetworks~\cite{hypersound} that generate INR weights conditioned on two modalities: an \textit{environment encoder}, extracting spatial priors from room images, and a \textit{pose encoder}, capturing human motion features. A generative model further augments realistic motion diversity, ensuring \sys produces semantically accurate, environment- and activity-specific mmWave signals.

Our INR-hypernetwork architecture generalizes across unseen environments while supporting applications such as data augmentation, signal super-resolution, and compact storage. To our knowledge, \sys is the first framework to unify continuous INR-based RF modeling with context-driven hypernetwork adaptation. Compared to prior work (e.g., RF-Genesis with 20–120s inference time), \sys synthesizes realistic, environment-specific signals in just 3.4s per activity, enabling efficient, real-time deployment without requiring additional data collection or user-specific retraining.

To demonstrate \sys's efficacy, we conduct extensive experiments using data from 10 indoor scenes and 12 human activities. The dataset includes synchronized mmWave radar signals, RGB-D images, and human pose sequences captured via motion capture systems. Our experiments show \sys achieves a complex SSIM of 0.88 and PSNR of 35 dB, outperforming existing methods. Using \sys-augmented datasets improves activity recognition accuracy by up to 7\% and reduces MPJPE in pose estimation by 15\%, demonstrating its effectiveness for mmWave signal generation and augmentation. Additionally, experiments on the publicly available HuPR~\cite{hupr} dataset show that \sys-synthesized and augmented data improves pose estimation performance by up to 10\%.

%% file: sections_revised/02_related.tex
\section{Related Work}

\parlabel{Implicit Neural Representations (INRs) and Hypernetworks:} INRs have emerged as a powerful framework for modeling high-dimensional data as continuous functions~\cite{inr, inr_compression, inras}. By parameterizing signals as neural networks, they enable efficient encoding, interpolation, and super-resolution. While widely applied in computer vision tasks such as image compression~\cite{inr_compression} and 3D reconstruction~\cite{inr_scene}, their use for spatiotemporal radar signals remains underexplored. Hypernetworks~\cite{hypernet}, which dynamically generate weights for other networks, have shown strong synergy with INRs in image~\cite{hyper_image} and audio domains~\cite{hypersound, hypersound}, but their potential for radar signal generation is only beginning to be explored.

\parlabel{Applications of mmWave Signals:} mmWave signals support tasks such as human pose estimation~\cite{mmMesh, pose1, hupr, pose3, pose4}, activity recognition~\cite{har1, har2}, and gesture recognition~\cite{drai}, but their performance often degrades in unseen environments due to RF noise. Few-shot adaptation methods alleviate this but require new data collection, while approaches like mmClip~\cite{mmclip} target unseen activities rather than environment variation. In contrast, \sys generates environment-specific synthetic signals, improving robustness of mmWave systems for both traditional models and few/zero-shot activity recognition.

\parlabel{mmWave Signal Generation:} RF-Genesis~\cite{rf_genesis} generates high-quality mmWave signals using simulation and ray tracing but requires 20 seconds to 2 minutes per signal, while RF-Diffusion~\cite{rfdiffusion} uses time-frequency diffusion but produces lower-quality signals. Neither approach addresses novel signal encoding, efficient representation, or super-resolution, and \sys is the first framework to explore these aspects.

%% file: sections_revised/04_method.tex
\section{Methodology}
\subsection{Signal Representation}
In this section, we present our methodology for extracting task-invariant and compressed signal representations using implicit neural representations~(INRs). First, we validate the ability of INRs to model individual instances of complex signals, such as mmWave radar data, with high fidelity. Next, we extend this framework by integrating hypernetworks with the INRs, enabling the development of a generalized representation capable of capturing the entire signal space across diverse environments and activities. 

\begin{figure}[t]
    \centering
    \begin{subfigure}[b]{1.25in}
        \centering
        
        \includegraphics[width=.9\textwidth]{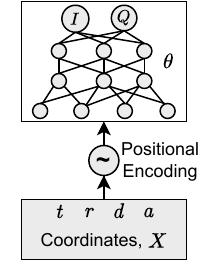}
        \caption{Base INR}
        \label{fig:inr_a}
    \end{subfigure}
    \hfill
    \begin{subfigure}[b]{1.85in}
        \centering
        
        \includegraphics[width=.9\textwidth]{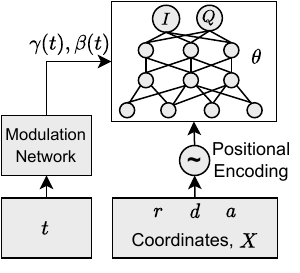}
        \caption{Proposed INR}
        \label{fig:inr_b}
    \end{subfigure}
    \caption{Implicit Neural Networks.} 
    %\vspace{-10pt}
    \label{fig:inr}   
\end{figure}

\subsubsection{INR for RF representation}  
\label{sec:inr}
\parlabel{Background:} Implicit Neural Representations~(INRs) are a powerful technique for modeling discretely sampled signals as continuous functions. An INR, \( f_\theta \), typically employs fully connected layers to map input coordinates \( \mathbf{X} \) to their corresponding signal values, enabling efficient encoding of high resolution data and supporting tasks like interpolation and super-resolution. For temporally varying signals, such as mmWave radar data, the input coordinates are extended to \( \mathbf{X} = (t, r, d, a) \), where \( t \) represents time, \( r \) denotes range, \( d \) indicates Doppler frequency, and \( a \) specifies the antenna index (azimuth or elevation). This basic formulation allows the INR to capture the spatio-temporal dynamics of mmWave signals, as shown in Figure~\ref{fig:inr_a} as \textit{Base INR}, enabling detailed and continuous modeling of such data.

\parlabel{Limitation:} The base architecture conflates spatial and temporal domains, making it inefficient for capturing temporal dynamics and requiring a larger number of parameters to model mmWave signals effectively. This inefficiency arises because, during human activity sensing, only a small subset of range or Doppler indices change between consecutive frames. As a result, using separate INR weights for each sparse radar frame becomes highly redundant and computationally inefficient.

\parlabel{Proposed Architecture:} To address the limitations, we propose an enhanced architecture that decouples temporal patterns from spatial features by introducing a dedicated temporal encoding module. This separation enables the model to better capture human motion dynamics, improving both efficiency and representation performance. In Figure~\ref{fig:inr_b}, we present our decoupled architecture that incorporates a temporal modulation network. This network maps temporal inputs \( t \) to modulation vectors \( \mathbf{\gamma}(t), \beta(t)\). The input to the INR is the coordinate tuple \( \mathbf{X} = (r, d, a) \), which is fed to the Fourier-based positional encoder~\cite{pe} layer, $PE$, which maps \(X\) to a higher-dimensional space using exponentially scaled frequencies. This encoding captures both low- and high-frequency variations, enabling the model to represent fine-grained and global details effectively:
$$PE(\mathbf{X}) = \left[ \cos(2^i \mathbf{X}), \sin(2^i \mathbf{X}) \right]_{i=0}^{L}$$

where \( L \) is the number of frequencies used. For our designed INRs, we set \(L=8\), which maps input coordinates \(\mathbf{X} \in \mathbb{R}^3\) to a positional encoding \(PE(\mathbf{X}) \in \mathbb{R}^{51}\), resulting in a 51-dimensional vector.

The positional encoding vectors are then passed through a series of MLP layers, which share common weights \(\theta\) across all frames in a sequence. The output of each layer \(L\) is modulated by a scale factor \(\gamma(t)\) and a shift factor \(\beta(t)\), both dependent on time \(t\). The network output is expressed as:
\[
f_\theta(\mathbf{X}; t) = \sigma_L \big( \mathbf{\gamma}_L(t) \odot (\mathbf{W}_L \sigma_{L-1}(\ldots) + \mathbf{b}_L) + \mathbf{\beta}_L(t) \big),
\]
where \(\sigma_L\) is the ReLU~\cite{relu} activation function at layer \(L\), \(\mathbf{W}_L\) and \(\mathbf{b}_L\) are the weight matrix and bias vector, and \(\odot\) denotes element-wise multiplication.

\parlabel{Sampling:} After computing the network output \(f_\theta(\mathbf{X}; t)\), we can sample the signal at each discrete value of \(\mathbf{X}\) to generate the complete mmWave signal. Sampling at finer intervals within the range of these indices enables higher-resolution data generation. For instance, let the original sampling intervals for range, Doppler, and azimuth be \(\Delta r, \Delta d, \Delta a\), respectively. By downscaling these intervals by factors \(n_r, n_d, n_a\), the resolution increases proportionally, and the total number of sampled points becomes:
\[
N_{\text{new}} = n_r \cdot n_d \cdot n_a \cdot N_{\text{orig}},
\]
where \(N_{\text{orig}}\) is the number of original discrete sampling points. This capability facilitates the generation of multi-resolution mmWave data. Additionally, sampling within a controlled error range around discrete coordinates allows for augmenting existing datasets, introducing realistic variability without compromising the data's integrity.

\subsubsection{Generalization via Hypernetworks}
As shown in Section~\ref{sec:inr}, INRs can represent discrete RF signals as continuous functions, enabling compression and super-resolution. However, training a separate INR for each signal instance is computationally expensive. The key challenge is to generalize signal representation across diverse environments and motions by mapping environmental and motion features to corresponding RF signals.

To this end, we leverage hypernetworks— meta-models that generate weights for target networks. Since RF propagation depends on both static environmental structures (e.g., walls, furniture) and dynamic human motion, we use two specialized encoders to extract relevant features. The hypernetwork uses these to generate INR weights \(\theta\), and modulation vectors \(\gamma(t)\), \(\beta(t)\), enabling signal generalization across scenes and activities with high fidelity.

\begin{figure}[t]
\centering
\includegraphics[width=\linewidth]{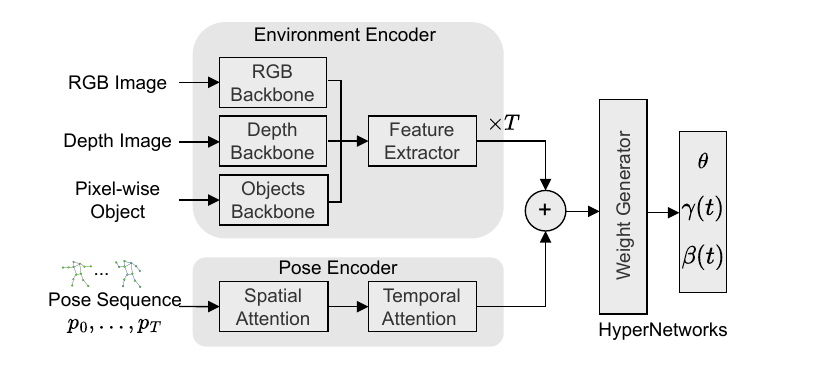}
\caption{Hypernetwork architecture with environment and pose encoders for generating INR weights.}
\label{fig:hypernetworks}
\vspace{-10pt}
\end{figure}

\parlabel{\textit{Environment Encoder}:} The encoder processes RGB images, depth maps, and object detection masks~\cite{kirillov2023segment} pretrained on COCO-Stuff~\cite{caesar2018coco}. A CNN based on EfficientNet-B0~\cite{tan2019efficientnet} extracts 512-dimensional spatial features \(\mathbf{z_e} \in \mathbb{R}^{512}\), balancing accuracy and efficiency across multimodal inputs.

\parlabel{\textit{Pose Encoder}:} Given 3D pose sequences \(\{\mathbf{p}_t \in \mathbb{R}^{J \times 3}\}_{t=1}^T\), a two-stage Transformer encoder captures spatial and temporal dependencies. Each pose \(\mathbf{p}_t\) is embedded via a fully connected layer: \(\mathbf{h}_t = \text{FC}(\mathbf{p}_t) \in \mathbb{R}^d\), then processed via spatial and temporal self-attention to yield a latent sequence \(\mathbf{z_p} \in \mathbb{R}^{T \times 512}\).

\parlabel{Hypernetwork:} The environmental encoding \(\mathbf{z_e}\) is broadcasted over time and concatenated with \(\mathbf{z_p}\). This joint representation is passed through MLP layers to generate the INR weights \(\theta\) and modulation vectors \(\gamma(t)\), \(\beta(t)\), which condition the INR to synthesize the radar signal at all spatial coordinates.

\begin{figure*}[t]
    \centering
    \begin{subfigure}[b]{5.5in}
        \centering
        \includegraphics[width=\textwidth,height=1in]{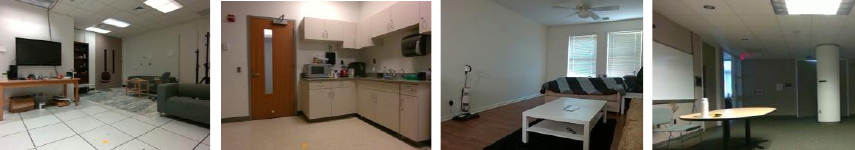}
        \caption{Synchronized mmWave and video data are recorded in diverse environments; four examples are shown.}
        \label{fig:envs}
    \end{subfigure}
    \hfill
    \begin{subfigure}[b]{1.2in}
        \centering
        \includegraphics[width=\textwidth, height=1in]{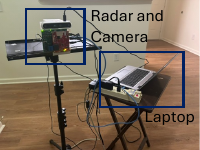}
        \caption{Device Setup.}
        \label{fig:setup}
    \end{subfigure}
    \caption{Examples of indoor environments and the device setup.}
    \label{fig:system}
\end{figure*}

\subsection{Signal Generation}

After training the INR and hypernetwork to map motion sequences to mmWave signals, we generate diverse synthetic data using MotionGPT~\cite{motiongpt}, a Transformer-based model for human motion synthesis. Given a motion prompt, MotionGPT produces temporally coherent 3D pose sequences by modeling spatial-temporal dependencies with multi-head self-attention. For each scene, the environment encoder extracts a feature vector \( \mathbf{z_e} \) from an RGB-D image, while MotionGPT-generated poses are encoded into latent motion vectors \(\mathbf{Z_p} = \{ \mathbf{z_p}(t) \}_{t=1}^{T}\). These are fused in the hypernetwork to generate INR parameters, which are sampled over \((r, d, a, t)\) to synthesize complex-valued mmWave signals. Repeating this process across varied prompts and environments yields diverse, realistic signals that augment training data.

%% file: sections_revised/05_microbenchmark.tex
\section{Experimental Setup}
\subsection{Dataset Collection}\label{data_setup}
We construct a diverse dataset to develop and evaluate our mmWave generation framework. mmWave data is captured using a TI AWR1843 radar~\cite{ti_awr1843} (77–81 GHz, 3 TX/4 RX, 4.4 cm range resolution), and visual data with an Intel RealSense D455 depth camera~\cite{intel_realsense_d455} ($1280\times720$). Both devices, housed in a custom case and connected to a Linux machine, are synchronized via timestamps (Figure~\ref{fig:setup}); the radar streams at 10 fps and the camera at 15 fps.

We collect data in 10 indoor scenes (offices, living rooms, labs, kitchens, libraries) with varied layouts and materials (Figure~\ref{fig:envs}). To increase diversity, we rearrange furniture, vary participant positions/orientations 5–10$\times$ per scene, and record six participants (ages 25–60, heights 5$^\prime$4$^{\prime\prime}$--6$^\prime$5$^{\prime\prime}$) performing 12 predefined activities (\textit{waving, clapping, jogging, jumping jacks, walking}, etc.), repeated multiple times. This yields over 12,000 frame sequences, providing a diverse dataset collected under our institution’s IRB protocol.

\subsection{Model Training}
The \textit{Signal Representation Module} learns a continuous function to model mmWave signals based on environmental features and human motion. It is trained using a combination of Complex SSIM Loss, Mean Squared Error (MSE) Loss, and Perceptual Loss. Perceptual Loss is an \(L_1\) loss applied to features extracted via a custom convolution-based mmWave feature extractor to capture high-level signal characteristics. Training runs for 1,000 epochs using the Adam optimizer with a learning rate of \(1 \times 10^{-4}\). At each epoch, the INR is sampled at 16,384 points from 131,072 spatial-temporal coordinates (\(r, d, a, t\)), prioritizing range indices with detected human activity while ensuring domain coverage through random sampling. The total loss function is:
\[
\mathcal{L}_{\text{INR}} = \lambda_{\text{SSIM}} \mathcal{L}_{\text{SSIM}} + \lambda_{\text{MSE}} \mathcal{L}_{\text{MSE}} + \lambda_{\text{Perceptual}} \mathcal{L}_{\text{Perceptual}},
\]
where \(\lambda_{\text{SSIM}}, \lambda_{\text{MSE}}, \lambda_{\text{Perceptual}}\) weight each term. We tuned $\lambda_{\text{SSIM}}{=}0.5$, $\lambda_{\text{MSE}}{=}0.3$, $\lambda_{\text{Perceptual}}{=}0.2$ on a validation split to balance fidelity and downstream utility. For a single signal instance, an INR is trained for 500 epochs with a learning rate of \(1 \times 10^{-4}\).

\begin{figure}[t]
\centering
\includegraphics[width=0.47\textwidth]{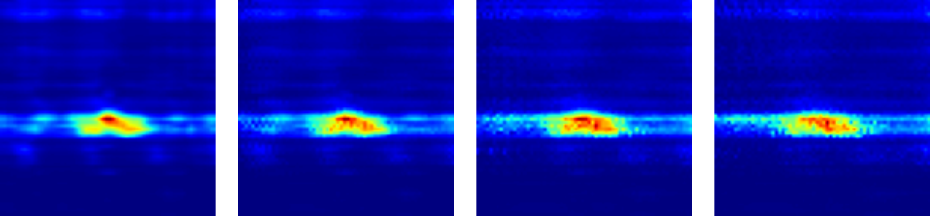}
%\vspace{-2em}
\caption{Range-azimuth spectrograms of generated signals at four sampling radii: 0, 1, 2, and 3~(from left to right).}
\label{fig:sampling_vis}
\vspace{-10pt}
\end{figure}

\subsection{Microbenchmarks}
\sys consists of an \textit{Environment Encoder} (2.1M parameters) for extracting spatial features from RGB-D images, a \textit{Pose Encoder} (4.9M) for capturing human motion, and a hypernetwork with supporting modules, totaling 14.73M trainable parameters. Each 20-frame sequence is represented by an INR with 10,018 parameters—4,898 for shared weights (\(\theta\)) and 5,120 for time-dependent modulations (\(\gamma(t), \beta(t)\)) across frames. Training and inference run on an NVIDIA RTX 4090 GPU (24 GB), where generating a 20-frame sequence takes ~3.4s, compared to 20s–2 min for RF-Genesis~\cite{rf_genesis}, highlighting \sys’s efficiency in high-resolution signal synthesis.

\subsection{Evaluation Metrics}
We evaluate \sys using a comprehensive set of metrics across signal quality, representation accuracy, and downstream performance.

\parlabel{SSIM~\cite{ssim} and cSSIM:}  
SSIM compares structural features such as luminance and contrast. For complex-valued mmWave signals, we use cSSIM, which separately evaluates the real and imaginary components to preserve both magnitude and phase information.

\parlabel{PSNR~\cite{psnr}:}  
PSNR measures reconstruction fidelity based on the ratio between the maximum signal value and the mean squared error, with higher values indicating better signal quality.

\parlabel{INR Evaluation:}  
To assess the INR's super-resolution and representation capabilities, we use cSSIM, MSE, and activity recognition accuracy. The latter is computed using a 12-class classifier trained on mmWave signals at different resolutions.

\parlabel{Point Cloud Evaluation:}  
We use Hausdorff distance~\cite{hausdroff} to measure geometric similarity between predicted and ground-truth point clouds, capturing maximum spatial deviation in 3D.

\parlabel{Downstream Task Evaluation:}  
To demonstrate real world utility, we fine-tune pretrained pose and action recognition models on \sys-generated signals in unseen environments, evaluating improvements in accuracy and robustness.

\begin{figure}[t]
\centering
\includegraphics[width=0.47\textwidth]{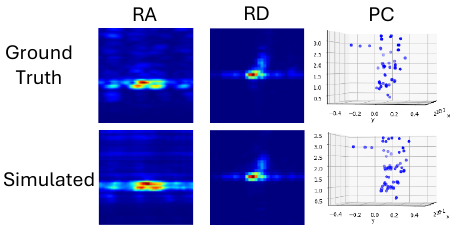}
%\vspace{-2em}
\caption{\sys-generated signal in range-azimuth, range-doppler spectrogram, and point cloud format.}
\label{fig:sample}
\vspace{-10pt}
\end{figure}

\subsection{Baselines}
To evaluate the effectiveness of data generation and augmentation, we utilize two baseline approaches for downstream tasks: mmMesh~\cite{mmMesh} for human pose estimation and the DRAI-based approach~\cite{drai} for human activity recognition. Additionally, we enhance the performance of these baseline models through environment-specific data augmentation provided by \sys, as well as two alternative data generation methods: RF-Genesis~\cite{rf_genesis}, which employs physical simulation for synthetic RF data, and the time-frequency diffusion-based approach RF-Diffusion~\cite{rfdiffusion}.

%% file: sections_revised/06_evaluation.tex
\begin{figure}[t]
    \centering
    \begin{subfigure}[b]{1.6in}
        \centering
        \includegraphics[width=\textwidth]{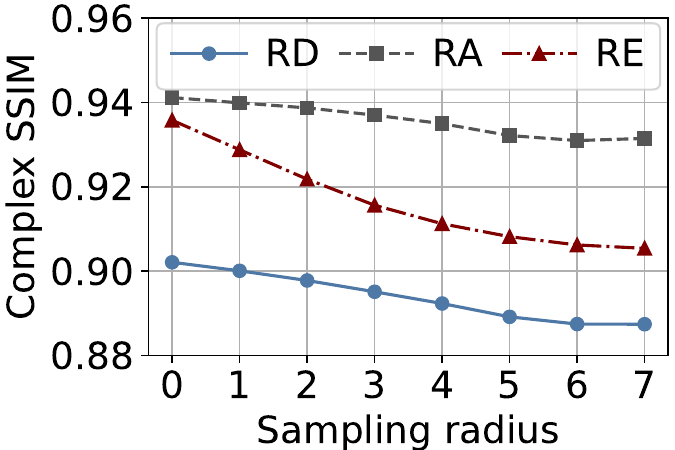}
        \label{fig:sampling_ssim}
    \end{subfigure}
    \hfill
    \begin{subfigure}[b]{1.6in}
        \centering
        \includegraphics[width=\textwidth]{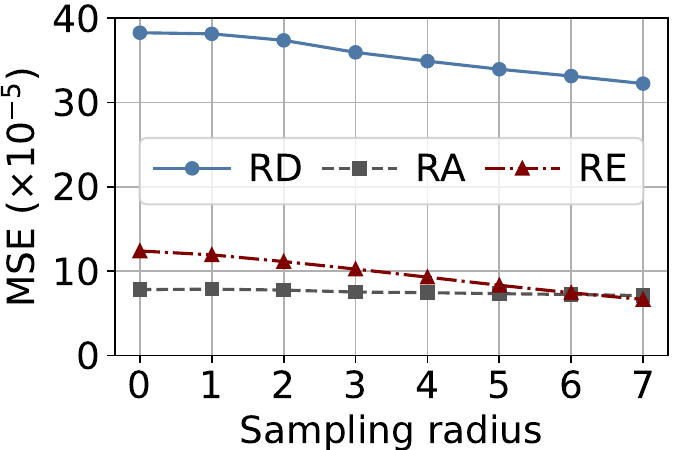}
        \label{fig:sampling_mse}
    \end{subfigure}
    \vspace{-10pt}
    \caption{Data augmentation using different sampling.}
    \label{fig:sampling_augmentation}    
\end{figure}

\begin{comment}
\begin{figure}[t]
\centering
\includegraphics[width=0.47\textwidth]{figs/random_sampling_vis.pdf}
%\vspace{-2em}
\caption{RD Spectrogram at different sampling.}
\label{fig:pose_est}
\end{figure}
\end{comment}

\begin{figure}[t]
    \centering
    \begin{subfigure}[b]{1.6in}
        \centering
        \includegraphics[width=\textwidth]{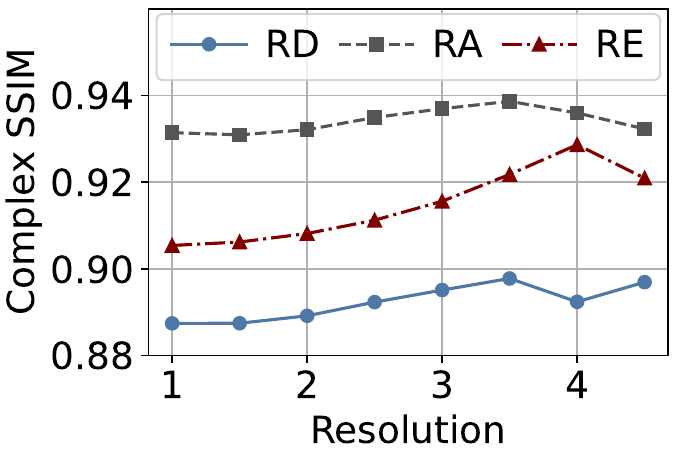}
        \label{fig:super_ssim}
    \end{subfigure}
    \hfill
    \begin{subfigure}[b]{1.6in}
        \centering
        \includegraphics[width=\textwidth]{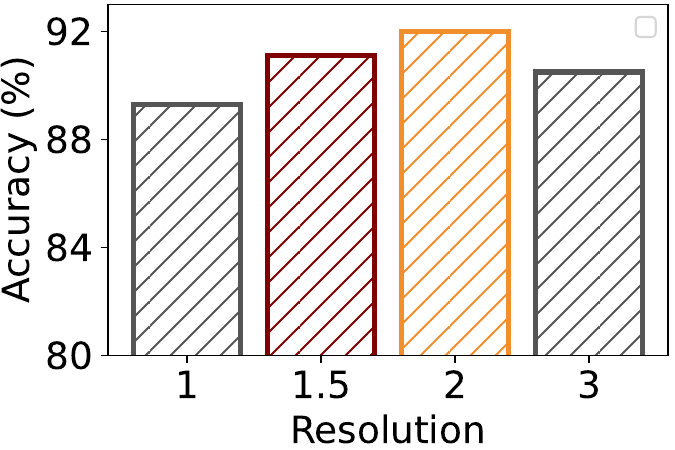}
        \label{fig:super_har}
    \end{subfigure}
    \vspace{-10pt}
    \caption{Benefits of higher resolution sampling using INR.}
    \label{fig:super_res}    
\end{figure}

\section{Evaluation}
\subsection{Evaluation of INR}
Using Implicit Neural Representations (INRs) to model continuous complex signals such as mmWave radar provides significant advantages beyond synthetic signal generation. In this section, we discuss the benefits of using INRs to augment and enhance existing mmWave radar data for downstream tasks.

\parlabel{Data Augmentation:} INRs provide a key advantage by learning a continuous function \( f_\theta \) over discretely sampled signals, enabling sampling at arbitrary coordinates. For example, given a radar signal with 64 discrete range indices, an INR trained on these indices can generate values at intermediate points (e.g., range index \(1.1\)), facilitating systematic data augmentation. This approach introduces realistic, controllable variations into the dataset, enhancing its diversity and robustness.

To evaluate augmentation, we divide index intervals into 8 parts and define a \textit{sampling radius} \( r \in [0,7] \). At coordinate \(j\), samples are drawn from \([j, j + \tfrac{r}{8}]\) and quality is measured via cSSIM and MSE on range-Doppler (RD), range-Azimuth (RA), and range-Elevation (RE) spectrograms (Figure~\ref{fig:sampling_augmentation}). For \(r \leq 2\), scores remain close to the original (\(r=0\)), with cSSIM dropping only 0.05–0.12 for RD/RA, while RE shows larger drops due to limited elevation resolution. Visualizations in Figure~\ref{fig:sampling_vis} confirm high-fidelity RA spectrograms at \(r=1,2,3\). These results demonstrate that INRs can augment mmWave datasets by varying spatial sampling, improving robustness of downstream models to signal variation and noise.

\parlabel{Super Resolution:} Another significant advantage of learning a continuous function over discretely sampled signals is the ability to perform super-resolution by sampling at augmented intermediate coordinates. For example, instead of sampling at discrete range indices \([0, 1, \dots, 64]\), we can sample at finer intervals, such as \([0, 0.5, 1, \dots, 64]\), effectively doubling the resolution of the original signal. This enables the generation of higher-resolution signals while preserving the underlying structure of the data. If the INR is well-fitted to the original signal, higher-resolution sampling enhances fine-grained information, which is expected to increase SSIM scores and improve downstream task performance.

\begin{figure}[t]
\centering
\includegraphics[width=0.47\textwidth]{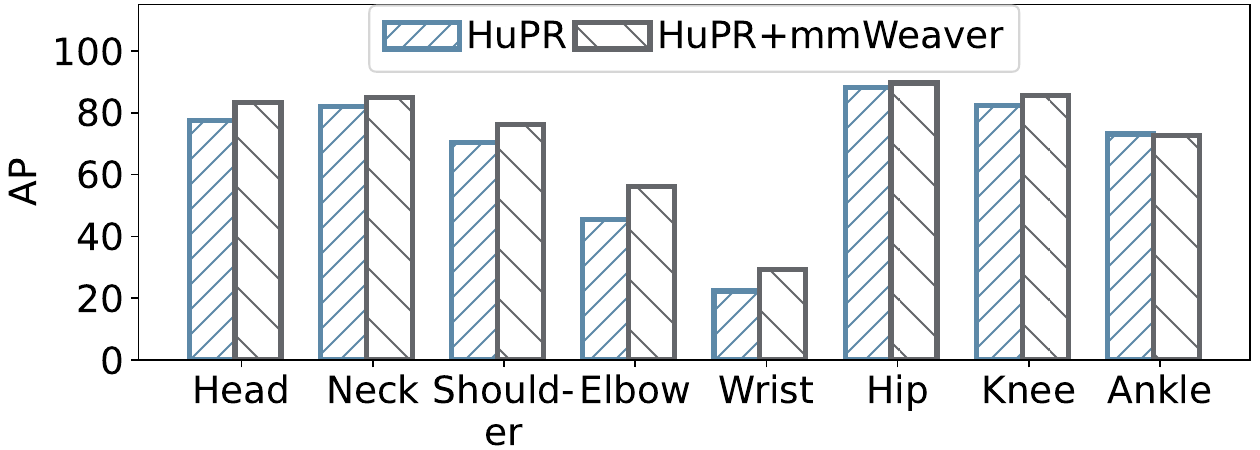}
%\vspace{-2em}
\caption{Joint-wise comparison on HuPR dataset.}
\label{fig:hupr_eval}
\vspace{-10pt}
\end{figure}

We evaluate the impact of super-resolution in Figure~\ref{fig:super_res}, where SSIM scores are reported for range-Doppler (RD), range-Azimuth (RA), and range-Elevation (RE) spectrograms at resolutions upto $ 4\times$ the original coordinates. The results demonstrate consistent improvements in SSIM scores across all spectrogram types with increasing resolution, indicating the fidelity of the super-resolved signals. Furthermore, we assess downstream task performance by training a baseline human activity recognition model on the original signal and high-resolution signals sampled at $1.5\times, 2\times$ and $3\times$ resolutions. The higher resolution signals yield $1-3\%$ improved recognition accuracy, showcasing the efficacy of super-resolution in enhancing both signal quality and downstream task outcomes.

\parlabel{Compression:} Our network uses $10{,}018$ parameters to model data of size $10 \times 64 \times 32 \times 12 \times 2 = 491{,}520$ points, yielding a compression ratio of $\tfrac{491{,}520}{10{,}018} \approx 49.06$.

\subsection{Evaluation of Signal Generation}
We evaluate \sys against RF-Genesis and RF-Diffusion using SSIM, PSNR, and Hausdorff Distance. SSIM and PSNR measure perceptual similarity and signal fidelity, respectively, while Hausdorff Distance quantifies geometric alignment between predicted and ground-truth point clouds. Higher SSIM/PSNR and lower Hausdorff values indicate better signal quality. Since text-conditioned generation yields multiple plausible outputs, we use controlled settings where synthetic signals are generated from exact 3D pose sequences for fair comparison.

\begin{figure}[t]
    \centering
    \begin{subfigure}[b]{1.6in}
        \centering
        \includegraphics[width=\textwidth]{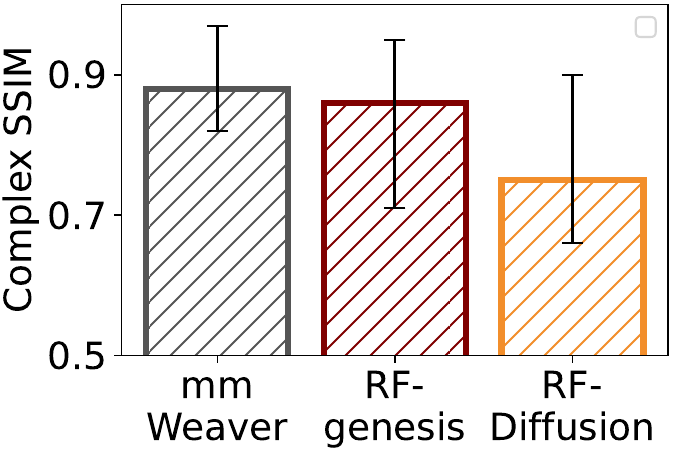}
        \label{fig:ssim}
    \end{subfigure}
    \hfill
    \begin{subfigure}[b]{1.6in}
        \centering
        \includegraphics[width=\textwidth]{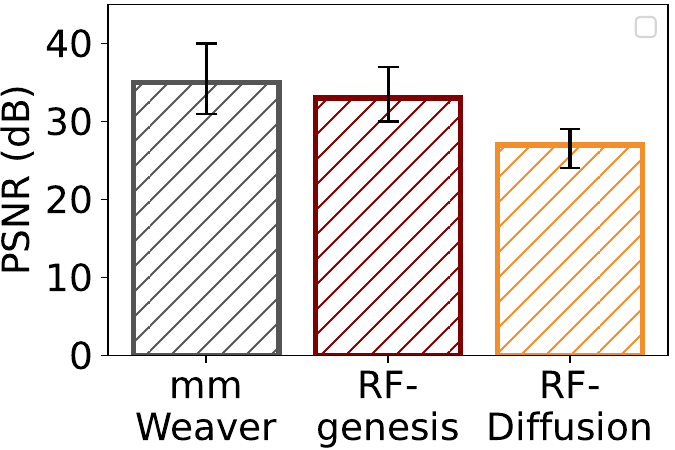}
        \label{fig:psnr}
    \end{subfigure}
    \caption{Comparison of quality of generated signal.}
    \label{fig:signal_eval}    
\end{figure}

As shown in Figure~\ref{fig:signal_eval}, \sys achieves the highest SSIM (\(0.88\)) and PSNR (\(35\ \text{dB}\)), outperforming RF-Genesis (\(0.86\), \(33.2\ \text{dB}\)) and RF-Diffusion (\(0.75\), \(31.6\ \text{dB}\)). For point cloud evaluation, \sys achieves the lowest Hausdorff Distance (\(0.32\ \text{m}\)) versus RF-Genesis (\(0.37\ \text{m}\)). Figure~\ref{fig:sample} shows sample RA/RD spectrograms and point clouds, illustrating \sys’s fidelity and structural accuracy. These results confirm \sys’s ability to synthesize perceptually and geometrically realistic mmWave signals.

\subsection{Generalization Across Unseen Environments}
We evaluate the robustness of \sys by categorizing three unseen test environments into clutter levels---\emph{low}, \emph{medium}, and \emph{high}---based on furniture density and occluders: a lab with minimal objects, a living room with standard furniture, and a small kitchen with dense clutter. Using the same set of activities, we augment training data with \sys-generated signals conditioned on the corresponding RGB-D priors and evaluate on unseen HAR tasks. In low clutter, we obtain cSSIM of $0.89$ and HAR accuracy of $90.6\%$; in medium clutter, cSSIM $0.87$ and HAR $91.0\%$; and in high clutter, cSSIM $0.84$ and HAR $86.7\%$. These results show that \sys preserves fidelity and recognition accuracy in simple and moderately complex scenes, with only modest degradation in highly cluttered scenes. Importantly, even in the most challenging case, \sys improves HAR performance ($86.7\%$) over the baseline ($83.2\%$), demonstrating robustness and consistent benefit.

\subsection{Generalization Across Unseen Activities}
We assess activity generalization by withholding two actions during training. For \emph{right-hand raise}, structurally similar to the trained \emph{left-hand raise}, \sys generated signals achieve $87.4\%$ accuracy, showing extrapolation to related motions. For the novel action \emph{jumping}, accuracy drops to $78.1\%$, reflecting the challenge of unseen activities. However, due to the modular design with separate encoders for activities and environments, \sys adapts quickly: fine-tuning on a small set of \emph{jumping} samples in one environment raises accuracy to $85.6\%$ across all environments. Thus, while zero-shot generalization to arbitrary activities is challenging, efficient adaptation is possible with minimal data.

\subsection{Evaluation on Public Dataset}
To assess \sys’s generalizability, we conduct experiments on the public HuPR~\cite{hupr} dataset and model for mmWave-based human pose estimation. HuPR contains 235 video sequences (60s each at 10 FPS), split into 193 training, 21 validation, and 21 test samples. As HuPR lacks explicit RGB-D scans of empty environments, we use the first video frame as a proxy. HuPR also uses two mmWave radars (horizontal and vertical), requiring two INRs per sequence to model dual perspectives.

To augment training, we map pose sequences from the evaluation set to mmWave signals via 1-to-1 mappings and apply structured random sampling for additional diversity. This increases variation in motion and context, improving generalization. We compare the baseline HuPR model trained on original data with HuPR+\sys, trained using both original and \sys-augmented data.

Figure~\ref{fig:hupr_eval} shows Average Precision (AP) across joints for both models. HuPR+\sys outperforms the baseline across all joints, with the largest gains in distal joints (elbow, wrist, knee) that are typically harder to localize due to occlusion and fast motion. These results validate the effectiveness of \sys-based signal generation for public dataset enhancement.

\begin{figure}[t]
\centering
\includegraphics[width=0.47\textwidth]{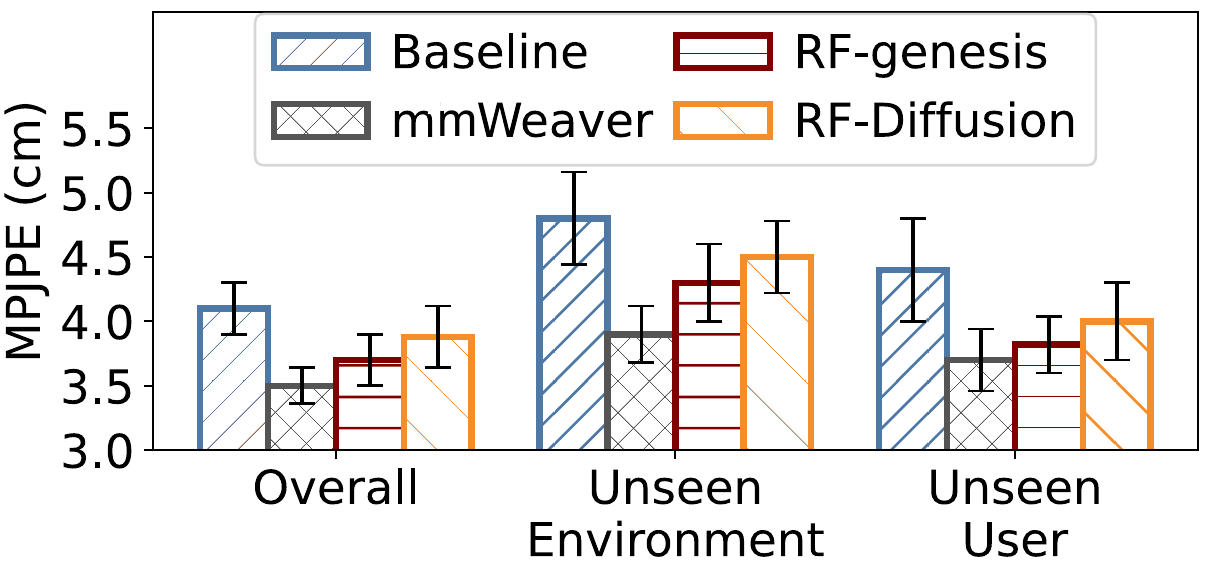}
\caption{Comparison of human pose estimation error.}
\vspace{-10pt}
\label{fig:pose_est}
\end{figure}

\section{Case Studies}
\subsection{Human Pose Estimation}
Human pose estimation is vital for applications like indoor monitoring and surveillance. In this section, we evaluate \sys’s effectiveness in enhancing pose estimation by augmenting existing mmWave datasets with high-quality, environment-specific synthetic data.

\subsubsection{Experiment Design:} 
We use pose data from 12 activities (Section~\ref{data_setup}) and generate 1,000 synthetic samples per method—RF-Genesis, RF-Diffusion, and \sys—to fine-tune a baseline pose estimation model. For \sys, samples are generated across all activities and environments for overall comparison. We further evaluate generalization in two scenarios: unseen users and unseen environments. In the latter, \sys generates synthetic data conditioned solely on RGB-D images and textual descriptions, requiring no real signal collection during testing. Additionally, we apply our INR-based data augmentation strategy during fine-tuning. This setup enables a comprehensive evaluation of \sys’s effectiveness across varied pose estimation scenarios.

\subsubsection{Overall Evaluation:}
For the overall evaluation, we allocate 80\% of the total data for training and reserve the remaining 20\% for testing. As shown in Figure~\ref{fig:pose_est}, the baseline model achieves an average pose estimation error of \(4.1\ \text{cm}\) in terms of MPJPE. When the dataset is augmented with synthetic data generated by \sys, RF-Genesis, and RF-Diffusion, followed by fine-tuning, the MPJPE decreases to \(3.5\ \text{cm}\), \(3.72\ \text{cm}\), and \(3.85\ \text{cm}\), respectively. These results highlight the effectiveness of dataset augmentation using synthetic data, particularly with \sys, which outperforms other methods. Additionally, the results emphasize the importance of combining new synthetic datasets with existing data augmentation techniques to improve pose estimation accuracy.

\subsubsection{Unseen Environment Evaluation:}
\label{sec:pose_unseen_env}
To evaluate performance in unseen environments, we reserve 2 of the 10 environments for testing and generate synthetic data using \sys. Results are compared against RF-Genesis and RF-Diffusion, which do not account for environment. As shown in Figure~\ref{fig:pose_est}, incorporating environment-specific synthetic data reduces MPJPE by \(1.0\ \text{cm}\), surpassing the \(0.6\ \text{cm}\) and \(0.4\ \text{cm}\) gains of RF-Genesis and RF-Diffusion. We further ablate environment priors by replacing exact priors with generic ones, yielding a \(0.7\ \text{cm}\) MPJPE reduction—comparable to environment-generic baselines but below the \(1.0\ \text{cm}\) achieved with \sys using exact priors.

\subsubsection{Unseen User Evaluation:}
Using a leave-one-out strategy across six users, \sys achieves \(3.7\,\text{cm}\) MPJPE compared to \(4.4\,\text{cm}\) baseline, outperforming RF-Genesis (\(3.8\,\text{cm}\)) and RF-Diffusion (\(3.9\,\text{cm}\)). While average MPJPE gains may appear modest, improvements of 0.6–1.1\,cm span 17 joints. Notably, some critical joints exhibit reductions up to \(10.45\,\text{cm}\), significantly enhancing downstream gesture and pose-based tasks.

\begin{figure}[t]
\centering
\includegraphics[width=0.47\textwidth]{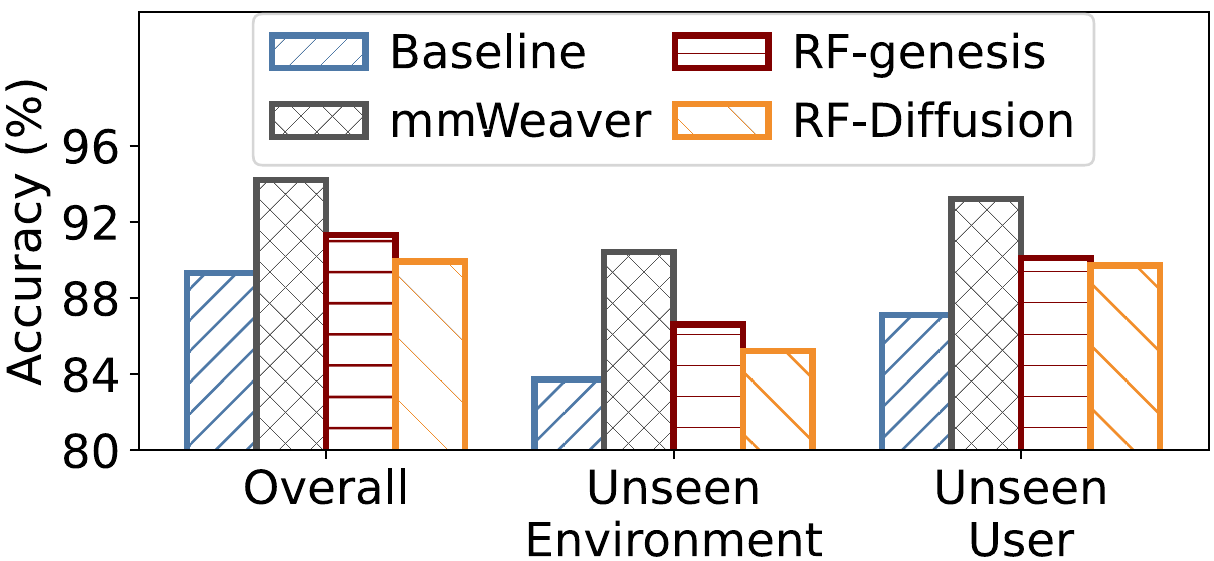}
%\vspace{-2em}
\caption{Comparison of human activity recognition.}
\vspace{-10pt}
\label{fig:har}
\end{figure}

\subsection{Human Activity Recognition}
Human activity recognition is critical for downstream tasks in applications such as indoor monitoring and behavioral analysis. In this section, we evaluate the capability of \sys to improve the performance of a baseline activity recognition model through synthetic data generation and augmentation.

\subsubsection{Experiment Design:} Similar to previous case study for human pose estimation, we evaluate activity recognition performance across three scenarios: overall performance, unseen environments, and unseen users. Synthetic datasets are generated using \sys, RF-Genesis, and RF-Diffusion, and the baseline model is fine-tuned with these augmented datasets. Classification accuracy is used as the evaluation metric, and results are reported in Figure~\ref{fig:har}.

\subsubsection{Results:} The baseline model achieves \(89.3\%\) accuracy, which rises to \(94.2\%\) with \sys-generated data. RF-Genesis and RF-Diffusion yield \(91.3\%\) and \(89.9\%\), respectively. For unseen environments, \sys reaches \(90.4\%\), outperforming RF-Genesis (\(86.6\%\)), RF-Diffusion (\(85.2\%\)), and the baseline (\(83.7\%\)). We also ablate environment priors by replacing exact priors with generic ones, yielding \(87.2\%\) accuracy—above the \(83.7\%\) baseline but below the \(90.4\%\) achieved with exact priors. For unseen users, \sys attains \(93.2\%\) accuracy, outperforming RF-Genesis (\(90.1\%\)) and RF-Diffusion (\(89.7\%\)), as well as the baseline (\(87.1\%\)).

%% file: sections_revised/07_conclusion.tex
\section{Conclusion}
We presented \sys, a framework that models mmWave signals as continuous functions using Implicit Neural Representations~(INRs), achieving $49\times$ compression and enabling multi-resolution sampling for data augmentation. To generalize across environments and activities, \sys employs hypernetworks conditioned on environmental and motion features, while a generative model synthesizes diverse motions to produce realistic, scenario-specific signals. Extensive evaluations show that \sys outperforms prior methods, improving both pose estimation and activity recognition. However, our system is currently evaluated on single-person scenarios; extending to multi-person cases is feasible (e.g., by superimposing signals of separated users), though handling close interactions remains an open challenge. Our setup also assumes a TI AWR1843 radar; while higher resolution can be approximated via augmented sampling, substantially higher-resolution radars would require hypernetwork fine-tuning. Future work will focus on optimizing INR efficiency, incorporating additional modalities and hardware, and broadening task applicability.